\documentclass[letterpaper, 10 pt, conference]{ieeeconf}  

\IEEEoverridecommandlockouts                              

\usepackage{graphics} 
\usepackage{amsmath} 
\usepackage{amssymb}  
\let\labelindent\relax

\DeclareMathOperator*{\argmax}{arg\,max}

\usepackage{epsfig}
\usepackage{caption}
\usepackage{svg}
\usepackage{multicol}
\usepackage{multirow}
\usepackage{import}
\usepackage{xcolor,colortbl}
\usepackage{enumitem}
\usepackage{cite}
\usepackage{hyperref}
\hypersetup{
    colorlinks=true,
    urlcolor=blue,
}

\usepackage{bbm}

\usepackage[symbol]{footmisc}

\title{\LARGE \bf
Crossing the Gap: A Deep Dive into  \\ Zero-Shot Sim-to-Real Transfer for Dynamics
}

\author{Eugene Valassakis$^{1}$, Zihan Ding$^{1}$, and Edward Johns$^{1}$
\thanks{$^{1}$The Robot Learning Lab at Imperial College London
        {\tt\small eugene.valassakis15@imperial.ac.uk}}%
}

\begin{document}

\maketitle
\thispagestyle{empty}
\pagestyle{empty}

\begin{abstract}

Zero-shot sim-to-real transfer of tasks with complex dynamics is a highly challenging and unsolved problem. A number of solutions have been proposed in recent years, but we have found that many works do not present a thorough evaluation in the real world, or underplay the significant engineering effort and task-specific fine tuning that is required to achieve the published results. In this paper, we dive deeper into the sim-to-real transfer challenge, investigate why this is such a difficult problem, and present objective evaluations of a number of transfer methods across a range of real-world tasks. Surprisingly, we found that a method which simply injects random forces into the simulation performs just as well as more complex methods, such as those which randomise the simulator's dynamics parameters, or adapt a policy online using recurrent network architectures.
\end{abstract}

\section{INTRODUCTION}

Zero-shot sim-to-real transfer aims to deploy a control policy trained in simulation to the real-world, without further training on task-specific real-world data. Much progress has been made in this field when learning vision-based policies with simple underlying dynamics, typically centred around the concept of domain randomisation, a technique that consists of injecting noise into different aspects of the simulation during training~\cite{tobin2017domain,james2017transferring}. However, transferring policies where the dynamics is complex is much more challenging, even when removing the need for vision transfer through the use of fiducial markers.

In our research, we originally set out to experimentally evaluate the field of zero-shot sim-to-real transfer for dynamics, aiming to determine the best methods for practical use. This revealed a somewhat surprising story that is not immediately obvious from reading previous published works. We found that works which have apparently shown great success, do not highlight enough the tedious engineering and task-specific tuning efforts required to achieve the published results, often simply stating the final simulator configurations used, which are not necessarily applicable to different tasks. Meanwhile, others have bypassed the real problem of interest by validating their methods only in simplified simulated environments, presenting results on sim-to-sim transfer rather than sim-to-real transfer. These combined shortcomings create a gap in the understanding between those with first-hand experience implementing these methods, and those reading such works and taking the results at face value.

\begin{figure}[t!]   
\centering    
\includegraphics[width=0.45\textwidth]{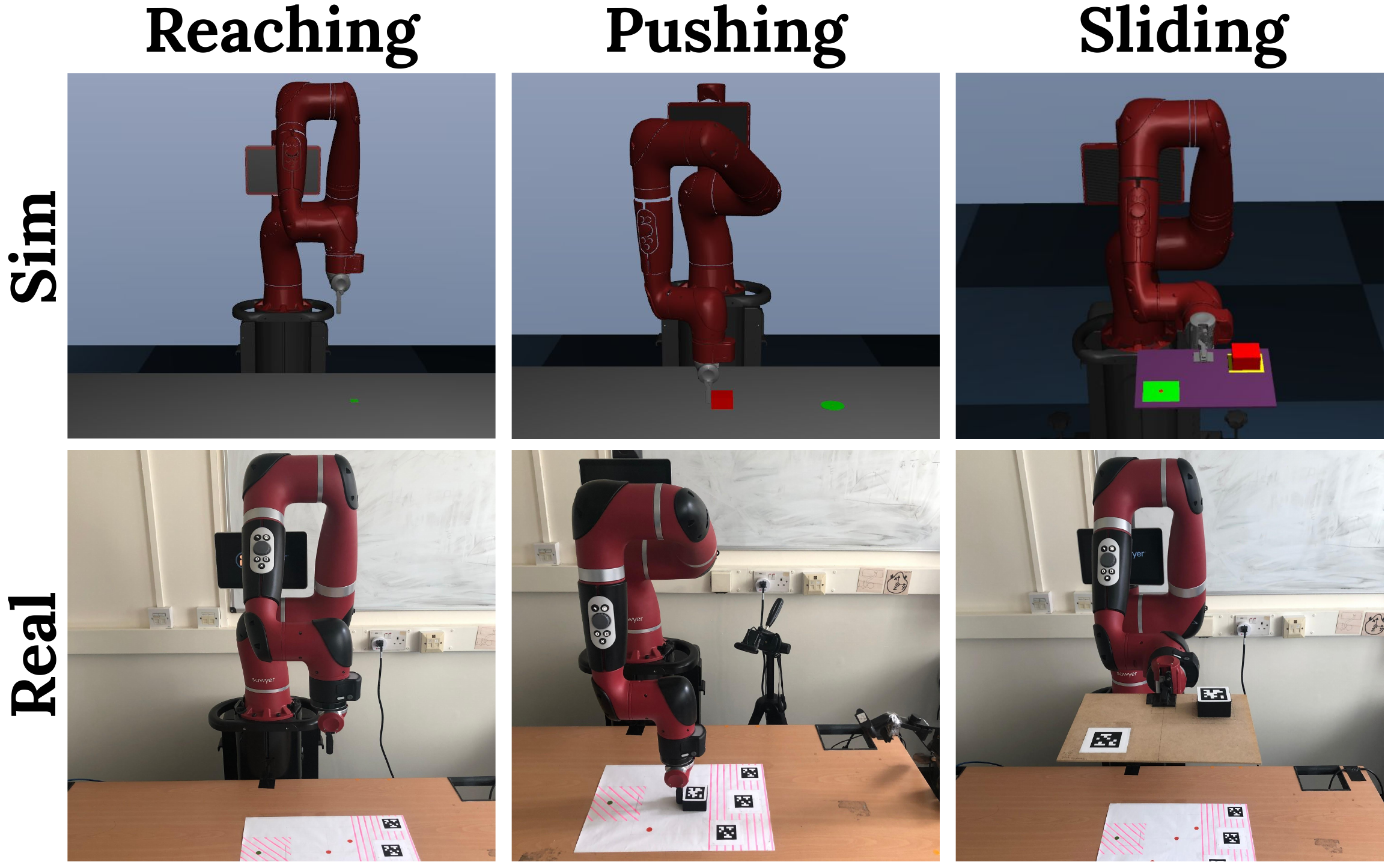}  
\vspace{-0.1cm}
\caption{Illustration of our experiment tasks in simulation and the real world.}
\label{fig:sim_real_setup}
\vspace{-0.6cm}
\end{figure}

In this paper, we aim to help the community cross that gap by providing a much more thorough examination of the problem than is typically seen in other works. We (i) shed some light on the fundamental aspects of sim-to-real transfer for a Reinforcement Learning (RL) policy, (ii) highlight the engineering process necessary to set up simulators for such a task, and (iii) provide thorough benchmarking and evaluation of recent methods, across three different tasks (see Figure~{\ref{fig:sim_real_setup}}), whilst only allowing for limited task-specific fine-tuning. In doing so, we also highlight some methods that we believe have been under-considered in the current literature, as well as provide real-world results for methods where published results exist only for sim-to-sim transfer. Through our experiments, we found that many of the more complex recent methods do not scale well to real-world tasks without significant task-specific tuning, which defeats the purpose of zero-shot transfer. We also found that some simple methods performed just as well, if not better, than more complex alternatives, whilst being significantly easier to implement and interpret. Most notably, we found that simply injecting random forces during simulation is at least as effective as the more typical approach of randomising the full range of simulation parameters. A video, supplementary material, and code, are available on our website\footnote[1]{\url{www.robot-learning.uk/crossing-the-gap}}.

\vspace{-0.1cm}
\section{RELATED WORK}\label{section:relwork}

Sim-to-real transfer can be seen as a special case of \emph{domain adaptation} (DA), the study of training models in a source domain to deploy them in a target domain. 
For  sim-to-real with RL in particular, methods typically address the visual~\cite{james2017transferring,tobin2017domain} or dynamics~\cite{peng2018sim, yu2017preparing,akkaya2019solving,andrychowicz2020learning,zhou2019environment,antonova2017reinforcement} adaptation problems, and either attempt zero-shot transfer~\cite{akkaya2019solving,james2017transferring}, or require real-world test-time data through meta-learning\cite{arndt2019meta}, residual learning~\cite{johannink2019residual}, real-world optimisation, ~\cite{yu2018policy} or mixed training between simulation and real-world~\cite{chebotar2019closing, buchler2020learning}. Our work focuses on zero-shot sim-to-real transfer for dynamics, and we review in more detail closely related works below. Although existing works have compared the accuracy of different physics simulators, in sim-to-sim \cite{erez2015simulation} or sim-to-real settings \cite{collins2019benchmarking}, to the best of our knowledge ours is the first to perform an extensive evaluation of sim-to-real transfer methods for dynamics.

\subsection{Dynamics Domain Randomisation}
Domain randomisation for dynamics sim-to-real transfer is a popular and widely adopted method, which involves randomising simulation parameters, such as physical parameters (e.g. mass and friction), control parameters (e.g. actuator gains), time delays, and observation noise. Although this idea has been around for a long time~\cite{jakobi1995noise}, it has recently been re-popularised with the advent of Deep Learning and Reinforcement learning, where it has shown successes in tasks such as object pushing~\cite{peng2018sim}, object pivoting~\cite{antonova2017reinforcement}, legged locomotion~\cite{hwangbo2019learning}, and tactile sensing~\cite{ding2020sim}. Perhaps one of the biggest success stories of this approach is OpenAI's work on dexterous manipulation~\cite{akkaya2019solving,andrychowicz2020learning}. Nonetheless, we believe these works underplay the amount of engineering effort required to achieve successful transfer, such as task-specific tuning of simulation parameters, which is not a scalable solution for zero-shot transfer.

As such, a number of works have proposed methods to automatically determine the distribution from which to sample these parameters~\cite{chebotar2019closing,mehta2019active,muratore2020bayesian, ramos2019bayessim,mozifian2019learning, akkaya2019solving}, most of which either require real-world data collection or have thus far only been evaluated on sim-to-sim transfer. Our work is orthogonal to these methods, as our findings suggest that there might be simpler parameter sampling spaces that lead to just as strong policy transfer, and any of the methods we benchmark could be augmented by such an automated process.

\subsection{System identification}
 System identification methods typically  consist of incorporating carefully identified properties and behaviours of the real environment into the simulators themselves~\cite{jeong2019modelling,allevato2020tunenet,tan2018sim, chebotar2019closing, chang2020sim2real2sim,kaspar2020sim2real, liang2020learning}. Our work focusses on the zero-shot regime, and Yu et al.~\cite{yu2017preparing}, demonstrate that system identification can be used online,  via training universal policies that can generalise across environments with different dynamics, and predicting the underlying dynamics parameters. This was extended in \cite{zhou2019environment} by conditioning policies on a latent representations of the dynamics. However, both these methods were only evaluated in sim-to-sim settings, and in this work we put them to the test in realistic sim-to-real scenarios.

\begin{figure*}[!tbh]
\centering
 \hspace{-0.1cm}
  \includegraphics[width=1.\linewidth]{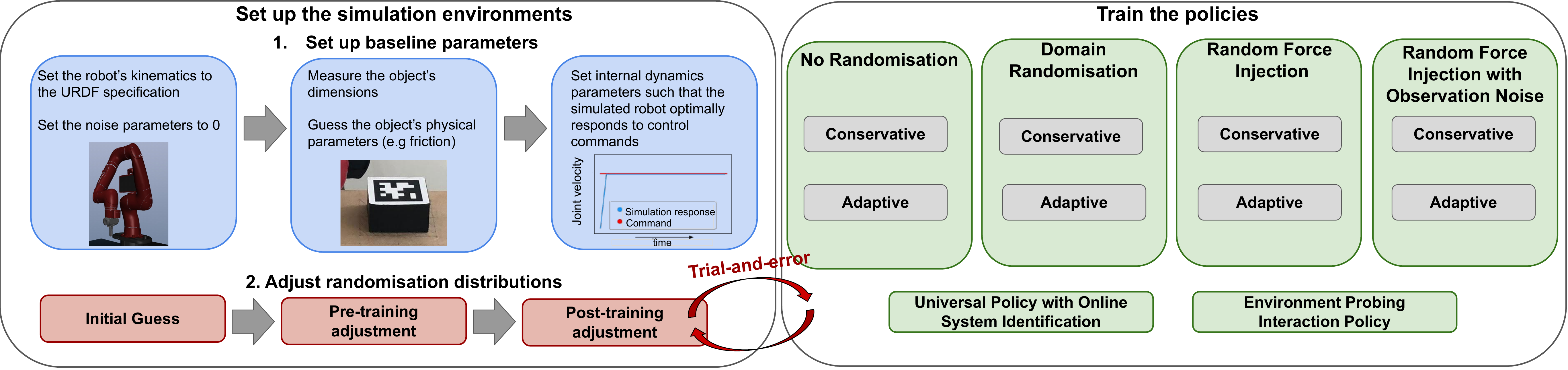}
  \caption{Block diagram of our methodology. On the left are depicted the steps followed to set up the simulator training environments, and on the right are the different combinations of  randomisation regimes and policy training methods tested. \label{fig:method_diagram}}
  \vspace{-0.7cm}
\end{figure*}

\section{MODELLING THE REALITY GAP} \label{section:reality_gap}
\subsection{Reinforcement Learning}

A robotics control task can be modelled as Partially Observable Markov Decision Process (POMDP), $\mathcal{M_{PO}}= \{ \mathcal{A}, \mathcal{S}, \mathcal{R}, \mathcal{P} , \gamma, \mathcal{O}, \mathcal{P_O}\}$, with $\mathcal{A}$,  $\mathcal{S}$, $\mathcal{O}$ the action, state and observation spaces, $\mathcal{R}$ the reward function, $\mathcal{P}$ and $\mathcal{P_O}$ the transition and emission probability distributions and  $\gamma$ the discount factor. At time $t$, given an action $a_t$, the state of the system $s_t$ transitions to $s_{t+1}\sim \mathcal{P}(s_{t+1}|s_t,a_t)$, and emits an observation $o_t \sim \mathcal{P_O}(o_t|s_t)$ and a reward  $r_t = \mathcal{R}(a_t,s_t)$. To solve the task, an RL algorithm finds a  parametric policy $\pi_{\theta}(a_t |o_t)$  that maximises the expected  return, $J(\theta)  = \mathbb{E}_{\theta}\big[{\sum_{t=1}^{\infty} \gamma^{t}r_{t}}\big]$.

\subsection{Sim-to-Real}

Under this POMDP model, the real world may differ from the simulator in the transition probability $\mathcal{P}$, and emission probability $\mathcal{P_O}$. Dealing with differences in $\mathcal{P}$ is referred to as dynamics sim-to-real transfer, and differences with $\mathcal{P_O}$ as visual sim-to-real transfer when the observations are images. The policy $\pi_{\theta}(a_t|o_t)$ typically involves inferring the underlying state $s_t$ (implicitly or explicitly) resulting in an effective policy $\mu(a_t|s_t)$. An important distinction between the dynamics and visual problems is that the policy $\mu$ should be adaptive to $\mathcal{P}$ and invariant to $\mathcal{P_O}$.

Domain randomisation is arguably the most popular way to deal with the sim-to-real transfer problem to date. It consists of sampling a set of simulation parameters $\mathbf{\xi} \sim p(\mathbf{\xi})$, creating an ensemble of training POMDPs which differ in $\mathcal{P}$ and/or $\mathcal{P_O}$. Its aim is to find a policy $\pi_{\theta^*}$ that can generalise across all these environments, formally $\theta^* = \argmax_{\theta}\mathbb{E}_{\xi}J_{\xi}(\theta)$, where $J_{\xi}$ is the objective for each individual environment. In practice, a new set of parameters $\mathbf{\xi}$ is drawn from $p(\mathbf{\xi})$ at the start of each training episode. The underlying hypothesis is that the real-world POMDP falls within the distribution of simulation POMDPs, allowing the simulation policy to transfer via interpolation.

In our work, we are mostly interested in the dynamics transfer problem, so we  minimise the impact of discrepancies in  $\mathcal{P_O}$ by using proprioceptive information and pose extractors to make low-dimensional observations for our policies (see Section~\ref{tasksdescription}). Nonetheless, effects such as observation noise, calibration errors and observation time delays are impossible to eliminate entirely. As such, although not further emphasising the distinction, we randomise both dynamics and observation parameters where appropriate.

\vspace{-0.1cm}
\section{METHODS}\label{section:methods}

In this section, we describe the framework we used to set up the simulator and implement the different sim-to-real transfer methods, which is illustrated in Fig.~\ref{fig:method_diagram}. We also describe the individual methods in detail, and distinguish between methods for randomising simulation parameters (\ref{dr_regimes}), and methods for training simulation policies using these parameter distributions (\ref{section:policies}).

\subsection{Environment Randomisation Regimes}\label{dr_regimes}

When creating a simulated environment for a real-world task, this environment is dependent upon a set of dynamics parameters. Choosing a distribution to sample these parameters from is a key challenge, and significant trial-and-error is typically required. A detailed breakdown of all simulator parameter values, distributions and ranges we used in our experiments can be found in our supplementary material, and here we describe the process we used in order to obtain those. In our implementations, we first find a set of \textit{baseline} parameters which result in reasonable simulator behaviour (blue blocks in Fig.~\ref{fig:method_diagram}):

\begin{itemize}
\item Kinematics parameters of the robot arm are set to the values described in the robot's URDF file. 
\item For other objects in the scene (the puck, the table, see Fig.~\ref{fig:sim_real_setup}), we set their geometric properties with physical measurements. Their physical parameters, such as friction, are notoriously hard to measure, and in our implementations we use educated guesses by looking up typical values for different materials, and visually inspecting for realistic behaviour on the simulator.
\item To optimise the dynamics parameters of the arm, such as controller gains, damping ratios and joint friction values, we use differential evolution~\cite{storn1997differential}, with a cost function that minimises the difference between some arbitrary control signal and the simulated robot's response (see left plot on Fig.~\ref{fig:sim2real_characterisation}).

\item Basic noise parameters, such as observation noise and time delays, are all set to 0.

\end{itemize} 

We now describe four randomisation regimes which further build upon these baseline values, each defining a different distribution of training environments for our policies.
 
\textbf{No Randomisation (NR)}.\label{norand}
Policies are trained using the baseline parameter values, with no randomisation.

\textbf{Domain Randomisation (DR)}.\label{fullrand}
Domain randomisation~\cite{peng2018sim, andrychowicz2020learning} consists of sampling sets of simulator parameters in order to create new environments at each episode of training. These include physical parameters such as masses and friction coefficients, actuation parameters such as controller gains, observation parameters such as time delays or measurement noise, and unmodelled effect parameters such as action noise. A more detailed description can be found in Section~\ref{tasksdescription}, with a full list in our supplementary material.  The baseline practice for zero-shot sim-to-real domain randomisation consists of following a tedious manual calibration and trial-and-error process in order to set the sampling distributions of these parameters such that the task can be solved. It is not always clear from previous works how this manual adjustment process is conducted, and in our methodology we decomposed the process into three key stages (red blocks on Fig.~\ref{fig:method_diagram}):

 \begin{figure}[t!]   
\centering    
\includegraphics[width=0.4\textwidth]{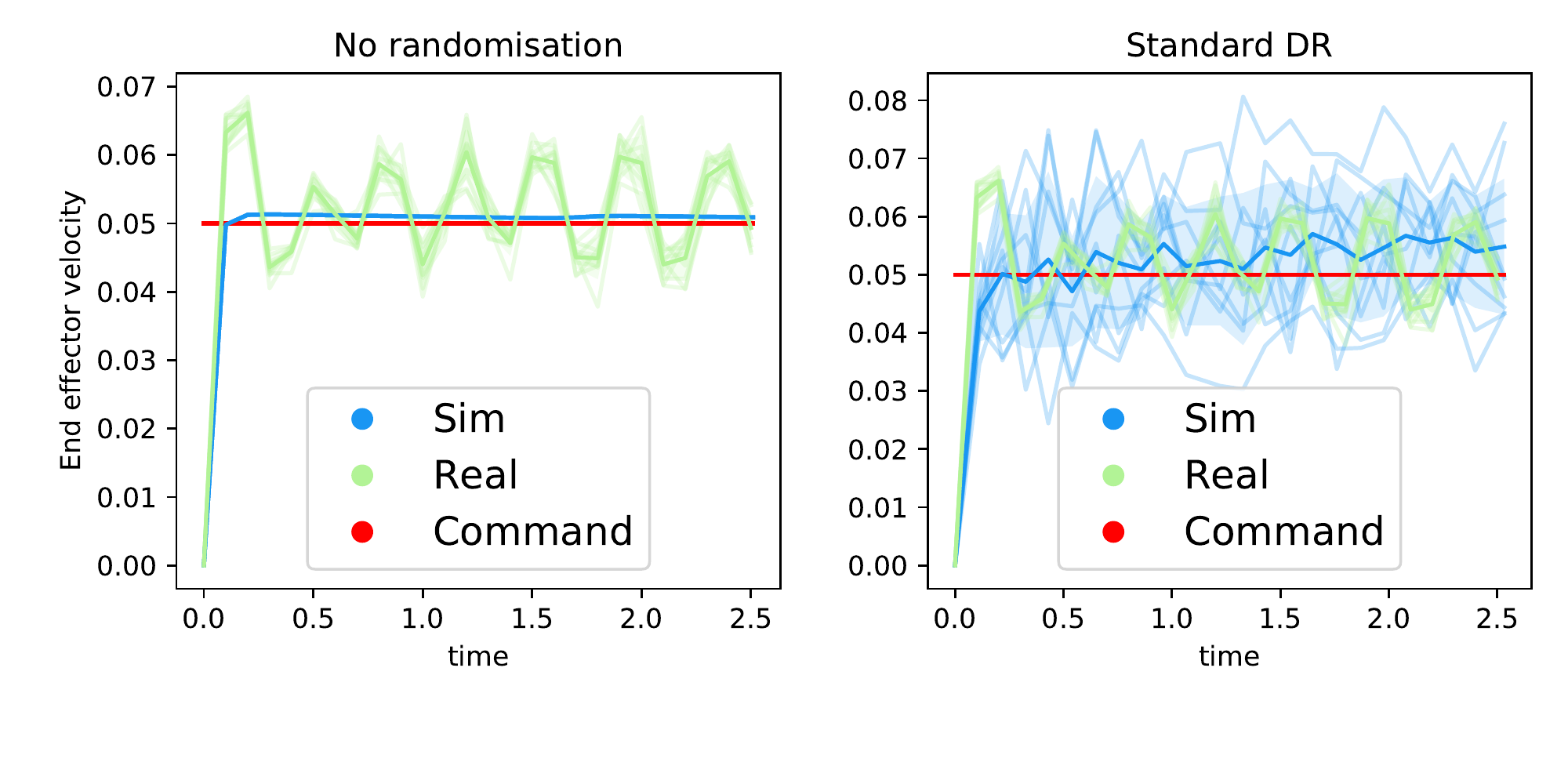}  
\vspace{-0.5cm}
\caption{Illustration of the reality gap (left) and the effect of randomising simulator parameters (right). For a constant velocity command, several real world trajectories and simulation trajectories are illustrated. Bold lines represent the mean. The y-axis corresponds to the y-component of end-effector velocity. The ``real'' and ``command'' curves are the same and repeated in all plots, for comparison. \label{fig:sim2real_characterisation} }  
\vspace{-0.55cm}
\end{figure}

\begin{enumerate}
    \item \label{initguess} \emph{Initial guess.}  We use a uniform distribution for parameters that are sampled from a fixed range, a loguniform distribution when sampling a multiplicative factor to the baseline value, and an equiprobable categorical distribution for discrete-value parameters. The initial range of these distributions is set using educated guesses, which relies on prior knowledge, domain expertise, and visual inspection of simulation behaviour.
    \item\label{pretraincal} \emph{Pre-training adjustment.} We create a hard-coded control policy for each task and execute it several times in the real world. We then execute the same policy in simulation the same number of times, using the parameter distributions defined in 1) to randomise the simulation. We plot the resulting trajectories, tracking the end-effector position $(x,y,z)$, as well as the position and orientation of objects involved in the task. One of these can be seen in the right diagram of Fig.~\ref{fig:sim2real_characterisation}, which illustrates the trajectory of the end-effector $y$ position in the reaching task. Visually inspecting these plots, we  adjust the ranges of the parameter distributions using trial-and-error until the trajectories obtained have similar variation to the real world trajectories.
    \item \emph{Post-training adjustment.} After training the task policies, we assess their performance both in simulation and the real world. If needed, we adjust the distributions, re-train the policies, and re-evaluate their performance. For a practical application, this would be repeated until the performance is satisfactory, relative to some real-world performance threshold for instance. In our experiments, we allowed for up to two post-training adjustments, in order to evaluate performance with minimal task-specific tuning. However, note that the policies themselves are never trained with real-world data.
\end{enumerate}

 In following this procedure, several challenges arise. First, specific parameter combinations may result in destructive simulator behaviours that greatly hurt policy learning. Second, the effect of varying one parameter is dependent on the values of all the remaining parameters which makes iterative adjustment challenging. Third, the dimensionality of the parameter space is vast, ranging between $31$ and $67$ depending on the task, and several parameters have esoteric interpretations, specifically dependent on modelling details of the simulator used.  In summary, adjusting these parameters requires heavy engineering efforts, significant domain-specific knowledge of the particular simulator and task, and typically post-training fine-tuning.

\textbf{Random Force Injection (RFI)}\label{section:rfi} Rather than randomising dynamics parameters, an alternative is to simply inject random noise into an environment with a fixed set of dynamics parameters. To the best of our knowledge, RFI has only briefly been considered in literature as a baseline to real-world data system identification methods~\cite{jeong2019modelling}, or as an unmodelled effect in addition to the full DR procedure~\cite{akkaya2019solving}. In this work, we examine RFI on its own merit, and show that this much simpler method can be just as effective as DR in crossing the reality gap.

A challenge in adding noise to simulator dynamics is the need to keep the visited simulation states stable, consistent, and physically realistic. This can be addressed by adding noise in the form of random generalised forces to the degrees of freedom of the simulator~\cite{jeong2019modelling}. Formally, RFI augments the generalised equations of motion of the simulated  rigid body system with  random generalised forces $\mathbf f_r \sim \mathcal{P}$ at each timestep, such that:  
\begin{align}
    \mathcal{M}(\mathbf{q}) \dot{\mathbf{v}} + c(\mathbf{q},\mathbf{v}) &= 
    \begin{bmatrix} \mathbf{0} \\ \mathbf{\tau} \end{bmatrix} +\sum_k^K\mathbf{J}_k^T(\mathbf{q})\mathbf{f}_k + \mathbf{f}_r,
\end{align}
with $\mathbf{q}$, $\mathbf{v}$ the generalised positions and velocities, $K$ the number of external forces $\mathbf f_k$, $\mathbf{J}_k^T$ the Jacobian mapping from external forces to generalised coordinates, $\mathbf{\tau}$ a torques vector, $\mathbf{0} $ a vector of zeros, $\mathcal{M}$ the mass matrix, and $c$ a bias forces vector~\cite{jeong2019modelling,Todorov2012MuJoCoAP}.

These forces inject white noise with a mean of 0, so the only parameters to set is the range of the uniform force sampling distribution for each dimension. Those can be adjusted using the same \textit{initial guess}, \textit{pre-training adjustment} and \textit{post-training adjustment} process described in the previous section. This is now a much simpler task than adjusting DR parameters, as (i) the dimensionality of the problem is much smaller, with only $7$ to $13$ dimensions to adjust depending on the task, and (ii)  the range of random uniform noise is conceptually simple with predictable effects, and therefore easier to tune than more complex dynamics parameters. Overall, this results in significantly reducing the amount of time and expertise required to optimise these parameters for a particular task. Intuitively, RFI aims to diversify the states observed in simulation such that this encompasses the real-world distribution of states, even though there is no explicit adaptation to the underlying dynamics itself as is possible with DR. However, our experiments showed that despite this, in practice RFI is at least as effective as DR, under the constraint of limited task-specific tuning.

\textbf{RFI with Observation Noise (RFI+)}.
Conceptually, as RFI only injects noise into the dynamics of the simulator, it only accounts for the reality gap in the transition distribution $\mathcal{P}$ of the POMDP underlying the task. As mentioned earlier, however, in real-world tasks it is impossible to completely eliminate the reality gap in the emission probability $\mathcal{P_O}$. In order to take into account these differences, we formulate the RFI+ training regime, which augments RFI by adding noise affecting the observation space of the policy, but not the underlying state of the world. This includes white noise in the observations, but also observation time delays. The sampling distributions for the parameters here are the same as their counterparts in DR and RFI. Overall, this adds another $2$ to $7$ parameters that need to be tuned, but maintains the desirable properties of RFI mentioned before.

\subsection{Transfer Policies}\label{section:policies}

From previous works, we have identified four main principles underlying zero-shot domain adaptation methods for dynamics~\cite{zhou2019environment, yu2017preparing, akkaya2019solving,andrychowicz2020learning,peng2018sim}. In this section we briefly describe these methods and our particular design decisions, with further implementation details in our supplementary material.

\textbf{Conservative Policy}.
A conservative policy receives a single environment observation input at time $t$ in order to compute the appropriate control action. Conceptually, this policy cannot infer information about the environment dynamics to help adjust its behaviour, as it does not have access to the effect of its actions on the state. As such, its optimal behaviour should be to take small, \emph{conservative} actions, such that if a large unexpected state transition occurs, the policy can still recover.

\textbf{Adaptive Policy}.
An adaptive policy has access to a sequence of observations and actions up to the current timestep $t$. Such a policy aims to create an internal representation of the current environment dynamics online, using the observed behaviour, and \emph{adapt} to these dynamics in real time. In order to implement the adaptive policy, we simply augment the core architecture with an LSTM layer branch as in~\cite{peng2018sim}.

\textbf{Universal Policy with Online System Identification (UPOSI)}.
With the UPOSI framework, Yu et al.~\cite{yu2017preparing} suggest to \emph{explicitly} represent the dynamics of an environment as the set of simulator parameters that define it, and regress those values with an \emph{online system identification} (OSI) module from a history of state-action inputs. Subsequently, these predicted parameters are input to a \emph{universal policy} (UP), alongside with the environment state, to enable online adaptation to dynamics.

In training UPOSI, we follow the procedure described in the original paper~\cite{yu2017preparing}.  We note that as opposed to~\cite{yu2017preparing}, we predict a much more diverse set of parameters, characterising dynamics of both the robot and external objects. In order to facilitate this prediction, our OSI network input is made by stacking 5 vectors  that include (i) the task-specific policy input states defined in Section~\ref{tasksdescription}, (ii) the policy actions, (iii) the robot's joint angles, (iv) the robot's joint velocities, and (v) the low-level joint velocity control derived from the actions.

\textbf{Environment Probing Interaction Policies (EPI)}.
Zhou et al.~\cite{zhou2019environment} proposed a framework similar to UPOSI, but with an alternative way of representing the environment dynamics. In their work, a probing policy interacts with the environment for a small number of timesteps, and the resulting state-action trajectory is passed in a dynamics embedding module that condenses it into a latent representation $z$. This value $z$ does not have a clear physical interpretation, but is trained in a way to distill the dynamics of the environment. In the zero-shot setting, the universal task policy picks up the task execution where the probing policy left off, and uses $z$ in its input to gain information about the environment's dynamics.

\section{EXPERIMENTS}\label{section:experiments}

\vspace{-0.1cm}
\subsection{Tasks}\label{tasksdescription}
In order to evaluate our implementations, we tested each on three different tasks of varying difficulties, Reaching, Pushing, and Sliding, which we now describe. Further experimental details can be found in the supplementary material.

All our simulation environments are created using the MuJoCo~\cite{Todorov2012MuJoCoAP} physics engine and the Robosuite~\cite{corl2018surreal} toolkit. All our experiments use a 7-DOF Sawyer robot arm with a main control loop at 10 Hz. The pushing and reaching task policies output end-effector Cartesian target velocities, converted to joint velocities through inverse kinematics, and the sliding task policies directly output target joint velocities. Object poses are extracted in the real world using  April tags~\cite{Wang2016}. For pushing, in order to get accurate measurements throughout the workspace, two cameras in different positions were used. All of our policy inputs are expressed in the end-effector frame unless otherwise specified.  All our tasks are finite-horizon, and do not terminate if the goal is reached before the end of the episode. For the sliding task, an episode may terminate early if the object falls off the sliding plate.

\begin{figure}[t!]   
\centering    
\includegraphics[width=0.45\textwidth]{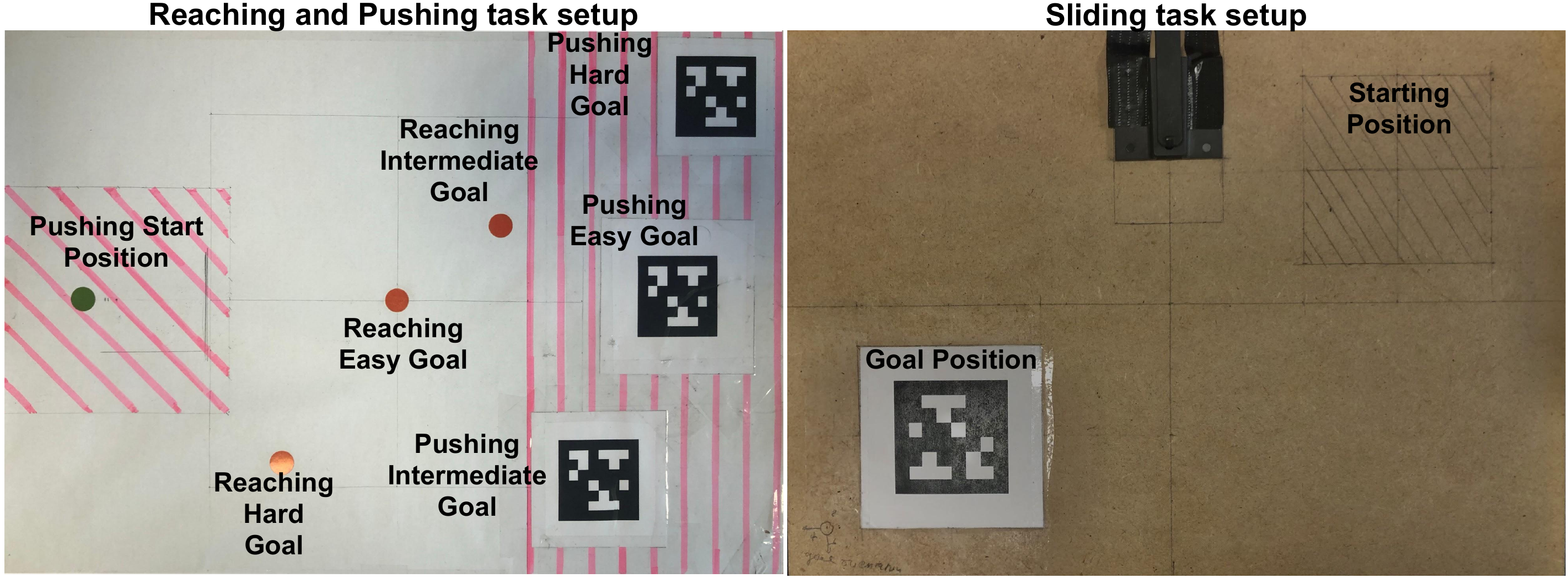}  
\caption{Start and goal positions for each task in the real world. Easy goals are well centred in the area of possible goals, hard goals are the furthest to the initial position, and intermediate goals are in the middle. For reaching, the end-effector begins directly above the easy goal.
\label{fig:real_world_task_setup} }   
\vspace{-0.6cm}
\end{figure}

\textbf{Reaching}.
In the reaching task, a target within an allowed region is specified on the table, and the robot has 6 s to reach a goal 2.5 cm above that target (see Fig.~\ref{fig:sim_real_setup}). Three goals (easy, intermediate, hard) are specified to test this task, as is shown in  Fig.~\ref{fig:real_world_task_setup}.

The policy inputs consist of the vector between the current position of the end-effector and the goal, concatenated with the 3D Cartesian velocity of the end-effector. In the real world, these are obtained using the proprioceptive sensors of the robot. The actions output by the policy form end-effector Cartesian velocities. The orientation of the end-effector constantly points vertically downwards. For domain randomisation, we randomise the control frequency, the PID feedback frequency, the PID gains, the observation noise variance, the action noise range, the robot link masses, and the joint damping, friction and armature values, giving a total of 56 parameters.

\begin{table*}[ht!]
\footnotesize
\centering
\caption{Main results for all experiments. Rates are averaged over all tested goals. The best performing method in the real world for each environment is in bold. The '-' indicates that the policy was too dangerous to run full experiments.\label{table:results}}
\centering
\addtolength{\leftskip} {-0.2cm} 
\setlength{\tabcolsep}{4pt}
\begin{tabular}{c|c|c|c|c|c|c|c|c|c}
\multicolumn{2}{c}{}  & \multicolumn{2}{|c|}{\textbf{Reaching}} &  \multicolumn{2}{|c|}{\textbf{Pushing}} & \multicolumn{4}{|c}{\textbf{Sliding}} \\
\multicolumn{2}{c}{} \vline  & \textbf{\begin{tabular}[c]{@{}c@{}}Sim \\Success Rate\end{tabular}} & 
\textbf{\begin{tabular}[c]{@{}c@{}}Real \\ Success Rate \end{tabular}} & \textbf{\begin{tabular}[c]{@{}c@{}}Sim \\Success Rate\end{tabular}} &
 \textbf{\begin{tabular}[c]{@{}c@{}}Real \\ Success Rate\end{tabular}}& \textbf{\begin{tabular}[c]{@{}c@{}}Sim \\Success Rate\end{tabular}} &
 \textbf{\begin{tabular}[c]{@{}c@{}}Real \\ Success Rate\end{tabular}} & \textbf{\begin{tabular}[c]{@{}c@{}}Sim \\ Fall Rate\end{tabular}} & 
 \textbf{\begin{tabular}[c]{@{}c@{}}Real \\ Fall Rate \end{tabular}} \\
  \hline
\multirow{2}{*}{\textbf{NR}} & \textbf{Conservative} & 1.0 &0.19  & 1.0 & 0.19 &1.0 &  0.0 & 0.0  &1.0\\
                            & \textbf{Adaptive}     & 1.0 & 0.0 & 0.67&  0.10 &0.94  & 0.0 & 0.0 & 1.0 \\
                                                                                       \hline
\multirow{2}{*}{\textbf{DR}}      & \textbf{Conservative} &0.99  & 0.62 &0.55   & 0.33& 0.4 &  0.0&0.08 &  0.0\\
                                      & \textbf{Adaptive} & 0.99 & 0.14  & 0.33 &0.33  & 0.68  & 0.0& 0.2&0.85 \\
                                                                                       \hline
\multirow{2}{*}{\textbf{RFI}}   & \textbf{Conservative} & 1.0 & \textbf{0.86}  &1.0  & \textbf{0.81} &0.78  &  \textbf{0.14}& 0.0 & 0.57\\
                                    & \textbf{Adaptive} & 1.0 & 0.10 & 0.33 & 0.38 & 0.8  & 0.0 & 0.04& 1.0\\
                                                                                       \hline
\multirow{2}{*}{\textbf{RFI+}}   & \textbf{Conservative} &0.99  & 0.71 &0.64  &0.38  & 0.56 & 0.0  &0.1 &0.14 \\
                                & \textbf{Adaptive}     & 0.99 & 0.29& 0.33 &0.14  & 0.44  & 0.0 &0.42&  0.43\\
                                                                                       \hline
\multicolumn{2}{c}{\textbf{UPOSI}} \vline& 0.0 & 0.0 & 0.0& -  & 0.22&0.0  &0.72 & 1.0\\
 \hline
\multicolumn{2}{c}{\textbf{EPI}}  \vline& 0.97  & 0.05 &  0.34 &0.38 & 0.0 & 0.0 & 1.0  &0.0

\end{tabular}
\vspace{-0.5cm}
\end{table*}

\textbf{Pushing}. The pushing task consists of pushing over 8 s a cuboid puck on the table, from a starting location to some specified goal within an allowed region (see Fig.~\ref{fig:sim_real_setup}). Again, at test time three goals are specified, as seen in Fig.~\ref{fig:real_world_task_setup}.

The plane of motion is defined by the table top, and the policy inputs consist of a concatenation of the in-plane components of (i) the vector between the end-effector and the puck, (ii) the vector between the puck and the goal, (iii) the end-effector and puck Cartesian velocities, and (iv) a sine-cosine encoding of the angle of rotation of the object around the surface normal of the table. The policy actions consist of in-plane end-effector Cartesian velocity commands, and the orientation of the end-effector is kept constant.

It is interesting to note that for pushing, the reality gap in observations is much stronger than for reaching, as pose information from fiducial markers (used for the puck) is much less accurate than the robot's proprioceptive sensors. Additionally to the parameters considered for reaching, domain randomisation here also includes time delays in the observations, object observation noise, and object physical properties such as mass, size and friction coefficients, for a total of $67$ parameters.

\textbf{Sliding}.
For the sliding task, we attach the end-effector of the robot to a rectangular cardboard plate, and designate a starting area and a specific goal position as seen in Fig.~\ref{fig:sim_real_setup} and Fig.~\ref{fig:real_world_task_setup}. The purpose of the robot is to tilt the sliding plate in a way to make the object slide  to the goal over $6$ s. As opposed to reaching and pushing, an episode may terminate early if the puck falls off the plate, and single goal location is used over both training and testing to facilitate learning.

In this task, we operate directly in joint-velocity space, and we only allow the  last two joints of the robot to move. The policy inputs consist of the object Cartesian position and velocity in the goal frame, the sine-cosine encoding of the rotation angle about the sliding plane normal, and the joint positions and velocities of the last two joints of the robot. The policy actions consist of joint velocity targets for the last two joints of the robot. The same parameters as with the pushing task are randomised, except that only the two relevant joints are considered.


\subsection{Network Architectures and RL Methods}
All our methods were fully trained in simulation using RL, with no further fine-tuning in the real world. For pushing and reaching, we randomised at each episode the starting positions of the end-effector and the goal within pre-specified regions, and for pushing we placed the puck next to the end-effector. For sliding, we randomised only the starting position of the puck within a specified region. In all tasks, the reward had a positive component for getting within a certain threshold of the goal, a negative reward component if the joint limits of the robot are hit, and a negative reward component consisting of the negative distance between the end-effector/puck and the goal. For reaching, the reward was augmented with a negative component for the end-effector hitting the table. For pushing, the reward was augmented with the negative distance between the end-effector and the object, and finally for sliding, it was augmented with a negative reward if the object falls off the sliding platform.

In order to make our benchmarking as fair as possible, we used a common core neural network architecture and the same reinforcement learning techniques for all methods. We used twin delayed deep deterministic policy gradients (TD3)~\cite{fujimoto2018addressing} with a fixed set of training hyperparameters as our main algorithm, and proximal policy optimisation (PPO)~\cite{schulman2017proximal} to train the probing policy in EPI. The network is fully-connected with 5 layers for the policy network and 4 layers for the Q-network, each with 512 units per layer and ReLU activations. The final layers  of the policy networks have Tanh activations  for normalising the action outputs. For UPOSI, the UP network follows our core architecture, and the OSI network has 4 fully connected hidden layers of $[512, 256, 128, 64]$ units, along with dropout layers. Deeper networks for the UP and longer sequences of motion state inputs to the OSI were considered, but neither seemed to improve results. For EPI, the dynamics embedding network has 3 hidden layers of 512 nodes and ReLU activations, and a Tanh activation for the output layer. Training $z$ also requires an extra prediction network, which in our case has 4 hidden layers with ReLU activations. We set $z$ to be $10$-dimensional, and allowed the probing policy to execute $10$ timesteps before the task-policy takes over.

\begin{figure*}[!tbh]
  \includegraphics[width=\linewidth,]{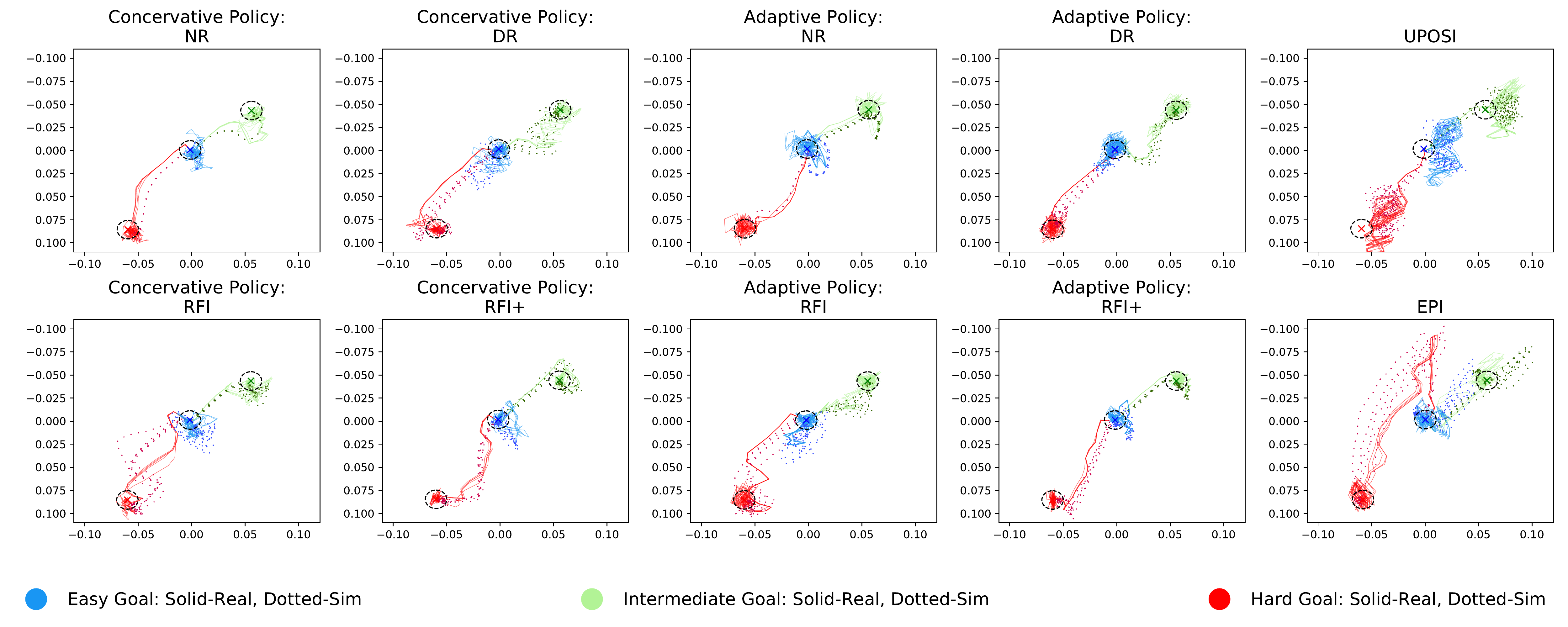}
  \vspace{-0.45cm}
  \caption{Horizontal trajectories of the end-effector over 5 trials on the reaching task. The axes distances are in meters.\label{fig:real_world_plot_results}}
  \vspace{-0.4cm}
\end{figure*}


\subsection{Experimental Setup}
For each of our tasks, as illustrated in Fig.~\ref{fig:method_diagram}, we trained both of the Conservative and Adaptive policies with each of NR, DR, RFI and RFI+, as well as the UPOSI and EPI  algorithms using the same environment randomisations as DR. In order to evaluate each method, we compare their performance in both simulation and reality, since some methods may find it difficult to learn even in simulation without extensive hyperparameter tuning. Therefore, we set up in simulation precisely the same experimental conditions (starting state, goal position), and deployed the same trained policies. In the real world, for each task and goal position considered, 7 trajectories were gathered, giving a total $469$ real world trajectories. In simulation, tests for each method were carried out with environments randomised the same way as during the training of that method, and $50$ trajectories were gathered for each task and goal giving a total of $3500$ trajectories, from which the average was calculated.

An episode was considered a success if the end-effector/object were within some threshold distance of the specified goal location over the final 0.5 s of execution. For the reaching task this threshold was 1 cm, for the pushing task it was 3 cm, and for the sliding task it was 2.3 cm. For the sliding task, we further recorded how often the object fell from the plate.

\subsection{Results}\label{section:results}

Table~\ref{table:results} shows a summary of our results, and a more detailed breakdown is in the supplementary material. Fig.~\ref{fig:real_world_plot_results} shows an illustration of the real-world trajectories for the reaching task. Examples of some trained policies are shown in our video.

From Table~\ref{table:results} it becomes apparent that RFI performed best across the board, and with most consistency. This is significant considering that DR took several days to tune, and we used post-training adjustment in an attempt to raise its performance (see Section~\ref{section:methods}), while the RFI distribution range was tuned within a few hours, and with no further task-specific post-training processing required. Nonetheless, when considering the full breakdown of results in our supplementary material, RFI did not always achieve the highest returns, and there was no strong overall winner. The trajectories plotted in Fig.~\ref{fig:real_world_plot_results} also show similarities in performances across all methods, except for NR and UPOSI which performed significantly worse.  We can, however, conclude that RFI overall performed no worse than other methods, whilst being significantly easier to implement, and significantly quicker to set simulation parameters for.
\begin{table}[t]
\footnotesize
\setlength{\tabcolsep}{1.5pt}
\centering
\caption{Comparing learned and noise inputs to the UP. Success rate across all goals for each task. When the success rate of both the learned and noise inputs are 0.0, the average return is shown in parenthesis. \label{table:universal_policies}}
\addtolength{\leftskip} {-0.1cm} 
\begin{tabular}{c|c|c|c|c|c|c}
\multicolumn{1}{c}{} &\multicolumn{2}{c}{\textbf{Reaching}} &\multicolumn{2}{c}{\textbf{Pushing}}  &\multicolumn{2}{c}{\textbf{Sliding}} \\ 
                   & \textbf{\begin{tabular}[c]{@{}c@{}}Learned\end{tabular}} &\textbf{\begin{tabular}[c]{@{}c@{}}Noise\end{tabular}}& \textbf{\begin{tabular}[c]{@{}c@{}}Learned\end{tabular}} &\textbf{\begin{tabular}[c]{@{}c@{}}Noise\end{tabular}} & \textbf{\begin{tabular}[c]{@{}c@{}}Learned\end{tabular}} &\textbf{\begin{tabular}[c]{@{}c@{}}Noise\end{tabular}}  \\
      \hline
\textbf{UPOSI }  & 0.0 (-4.8)  & 0.0 (-10.7)    &    0.0 (-23.2)  &    0.0 (-23.2) &      0.34   &    0.0     \\
\textbf{EPI} &     0.95       &     0.02  &     0.33    &    0.033    &     0.0 (-17.6)     &  0.0 (-40.9)   
\end{tabular}
 \vspace{-0.5cm} 
\end{table}

%

We further note that RFI generally performed better than RFI+, which is surprising given that RFI+ conceptually accounts more thoroughly for the sim-to-real discrepancies, and we conjecture that RFI+ could benefit from further tuning of the parameter ranges. Nonetheless, an interesting observation is that for the sliding task, RFI+ resulted in a more stable policy, rarely falling off the platform but also unable to reach the goal, while RFI resulted in a much more aggressive policy, often falling off the platform but sometimes managing to successfully complete the task. Overall, this highlights a trend that seems to form: the more parameters that are considered for environment randomisation, the more tuning of those parameters that needs to be done for the performance to be satisfactory, while simpler alternatives still seem to work just as well. 

Finally, an important general trend is that our adaptive policies performed worse than our conservative policies. As such, they seem to have failed in capturing and exploiting environment dynamics information, which is contrary to what some previous works suggest~\cite{akkaya2019solving}. A possible explanation for this is that, due to the short episode length, the adaptive policy does not have enough time to extract meaningful information about each environment before the environment resets. It is also possible that further careful tuning of our network and training hyperparameters might have allowed us to observe these properties. At the very least, we can conclude that adaptive policies are not always preferable to conservative policies as other works seem to suggest, because they appear to require stronger engineering efforts and fine tuning to lift their performance above a conservative policy.




\begin{figure}[t]   
\centering    
\includegraphics[width=0.3\textwidth]{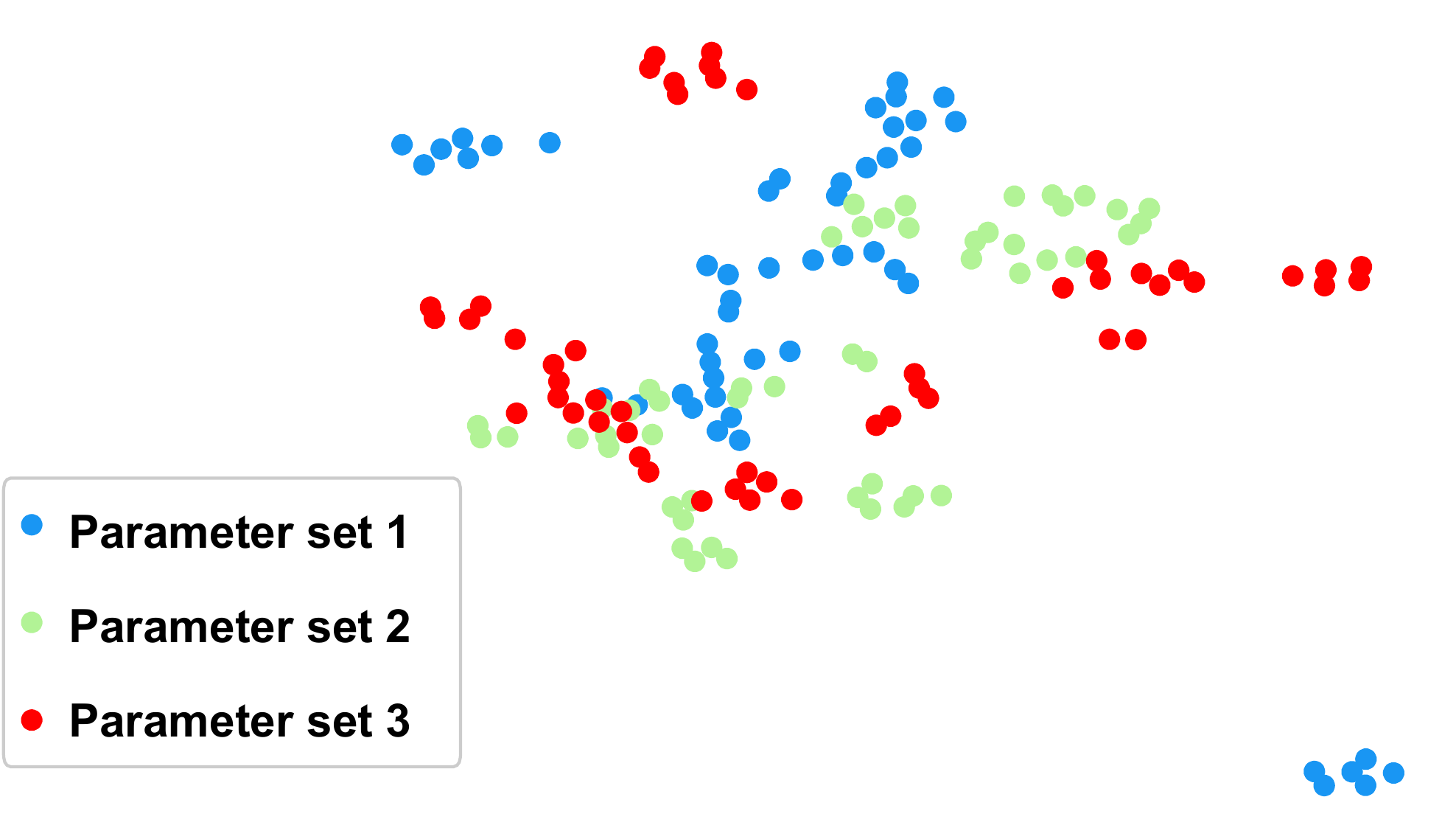}
\vspace{-0.2cm}
\caption{t-SNE plot of EPI latent vectors $z$ obtained for three distinct sets of environment dynamics.  \label{fig:tsne} }  
\vspace{-0.1cm}
\end{figure}

\begin{table}[t]
\centering
\caption{Error in the OSI prediction for each task.\label{table:osi}}
\begin{tabular}{c|c|c}
  \textbf{Reaching}    & \textbf{Pushing }& \textbf{Sliding}\\\hline
10.9~\% &   11.0~\%   &    12.7~\%  
\end{tabular}
\vspace{-0.3cm}
\end{table}

\subsection{Investigating System Identification}

To better understand the UPOSI and EPI results, we tested their universal policy and system identification components in isolation. For testing the universal policies, we show in Table~\ref{table:universal_policies} the effect of using white noise as input to the UP, rather than the predicted parameters. In order to test the OSI component of UPOSI, we show in Table~\ref{table:osi} the average \%~error in the predictions of OSI across our simulated tasks, relative to the range of the OSI predictions. We can not do this with EPI, so instead we take the following approach: we take three very different sets of parameters for the reaching task, and for each roll out 50 trajectories. We then show in Fig~\ref{fig:tsne} a two dimensional t-SNE~\cite{vanDerMaaten2008} graph of the latent variables z obtained for each of the three parameter sets. 

As we can see from Table~\ref{table:universal_policies}, with the exception of UPOSI on the pushing task, both universal policies are utilising the system identification module, since this performs significantly better than a policy conditioned on white noise. However, Table~\ref{table:osi} and Fig~\ref{fig:tsne} show that the system identification modules are overall not very accurate. In the case of EPI for instance, we see that although the three parameter sets tested are quite different, the latent space does not differentiate well between them. This suggests that inferring dynamics information from state-action sequences is a difficult task, and in fact ill-posed, as different parameter combinations may result in the same observed behaviour.

\section{CONCLUSIONS}
In this study, we have thoroughly examined the fundamental aspects and engineering processes involved in zero-shot sim-to-real transfer for dynamics. We compared, across three tasks, several methods which previously either have not been given significant exposure, or have been evaluated only on simplified environments. Amongst these, we have not found a clear winner, although it appears that all methods relying on inferring environment dynamics online trail behind. Overall, we believe that an important, and somewhat surprising, takeaway from our results is that, for zero-shot transfer where the amount of task-specific fine-tuning is limited, the standard practice of domain randomisation performed no better than simple random force injection. This is significant, because random force injection is much more practical to set up than full dynamics parameter randomisation, and dramatically lowers the engineering efforts and expertise required to achieve sim-to-real transfer.

\addtolength{\textheight}{-12cm}   






\bibliography{references}

\begin{thebibliography}{10}

\bibitem{akkaya2019solving}
I.~Akkaya, M.~Andrychowicz, M.~Chociej, M.~Litwin, B.~McGrew, A.~Petron,
  A.~Paino, M.~Plappert, G.~Powell, R.~Ribas, et~al.
\newblock Solving rubik's cube with a robot hand.
\newblock {\em arXiv preprint arXiv:1910.07113}, 2019.

\bibitem{allevato2020tunenet}
A.~Allevato, E.~S. Short, M.~Pryor, and A.~Thomaz.
\newblock Tunenet: One-shot residual tuning for system identification and
  sim-to-real robot task transfer.
\newblock In {\em Conference on Robot Learning}, pages 445--455, 2020.

\bibitem{andrychowicz2020learning}
O.~M. Andrychowicz, B.~Baker, M.~Chociej, R.~Jozefowicz, B.~McGrew,
  J.~Pachocki, A.~Petron, M.~Plappert, G.~Powell, A.~Ray, et~al.
\newblock Learning dexterous in-hand manipulation.
\newblock {\em The International Journal of Robotics Research}, 39(1):3--20,
  2020.

\bibitem{antonova2017reinforcement}
R.~Antonova, S.~Cruciani, C.~Smith, and D.~Kragic.
\newblock Reinforcement learning for pivoting task.
\newblock {\em arXiv preprint arXiv:1703.00472}, 2017.

\bibitem{arndt2019meta}
K.~Arndt, M.~Hazara, A.~Ghadirzadeh, and V.~Kyrki.
\newblock Meta reinforcement learning for sim-to-real domain adaptation.
\newblock {\em arXiv preprint arXiv:1909.12906}, 2019.

\bibitem{buchler2020learning}
D.~B{\"u}chler, S.~Guist, R.~Calandra, V.~Berenz, B.~Sch{\"o}lkopf, and
  J.~Peters.
\newblock Learning to play table tennis from scratch using muscular robots.
\newblock {\em arXiv preprint arXiv:2006.05935}, 2020.

\bibitem{chang2020sim2real2sim}
P.~Chang and T.~Padir.
\newblock Sim2real2sim: Bridging the gap between simulation and real-world in
  flexible object manipulation.
\newblock {\em CoRR}, abs/2002.02538, 2020.

\bibitem{chebotar2019closing}
Y.~Chebotar, A.~Handa, V.~Makoviychuk, M.~Macklin, J.~Issac, N.~Ratliff, and
  D.~Fox.
\newblock Closing the sim-to-real loop: Adapting simulation randomization with
  real world experience.
\newblock In {\em 2019 International Conference on Robotics and Automation
  (ICRA)}, pages 8973--8979. IEEE, 2019.

\bibitem{collins2019benchmarking}
J.~Collins, J.~McVicar, D.~Wedlock, R.~Brown, D.~Howard, and J.~Leitner.
\newblock Benchmarking simulated robotic manipulation through a real world
  dataset.
\newblock {\em IEEE Robotics and Automation Letters}, 5(1):250--257, 2019.

\bibitem{ding2020sim}
Z.~Ding, N.~Lepora, and E.~Johns.
\newblock Sim-to-real transfer for optical tactile sensing.
\newblock In {\em IEEE International Conference on Robotics and Automation
  (ICRA)}, 2020.

\bibitem{erez2015simulation}
T.~Erez, Y.~Tassa, and E.~Todorov.
\newblock Simulation tools for model-based robotics: Comparison of bullet,
  havok, mujoco, ode and physx.
\newblock In {\em 2015 IEEE international conference on robotics and automation
  (ICRA)}, pages 4397--4404. IEEE, 2015.

\bibitem{corl2018surreal}
L.~Fan, Y.~Zhu, J.~Zhu, Z.~Liu, O.~Zeng, A.~Gupta, J.~Creus-Costa, S.~Savarese,
  and L.~Fei-Fei.
\newblock Surreal: Open-source reinforcement learning framework and robot
  manipulation benchmark.
\newblock In {\em Conference on Robot Learning}, 2018.

\bibitem{fujimoto2018addressing}
S.~Fujimoto, H.~Van~Hoof, and D.~Meger.
\newblock Addressing function approximation error in actor-critic methods.
\newblock {\em arXiv preprint arXiv:1802.09477}, 2018.

\bibitem{hwangbo2019learning}
J.~Hwangbo, J.~Lee, A.~Dosovitskiy, D.~Bellicoso, V.~Tsounis, V.~Koltun, and
  M.~Hutter.
\newblock Learning agile and dynamic motor skills for legged robots.
\newblock {\em Science Robotics}, 4(26):eaau5872, 2019.

\bibitem{jakobi1995noise}
N.~Jakobi, P.~Husbands, and I.~Harvey.
\newblock Noise and the reality gap: The use of simulation in evolutionary
  robotics.
\newblock In {\em European Conference on Artificial Life}, pages 704--720.
  Springer, 1995.

\bibitem{james2017transferring}
S.~James, A.~J. Davison, and E.~Johns.
\newblock Transferring end-to-end visuomotor control from simulation to real
  world for a multi-stage task.
\newblock In {\em CoRL}, 2017.

\bibitem{jeong2019modelling}
R.~Jeong, J.~Kay, F.~Romano, T.~Lampe, T.~Rothorl, A.~Abdolmaleki, T.~Erez,
  Y.~Tassa, and F.~Nori.
\newblock Modelling generalized forces with reinforcement learning for
  sim-to-real transfer.
\newblock {\em arXiv preprint arXiv:1910.09471}, 2019.

\bibitem{johannink2019residual}
T.~Johannink, S.~Bahl, A.~Nair, J.~Luo, A.~Kumar, M.~Loskyll, J.~A. Ojea,
  E.~Solowjow, and S.~Levine.
\newblock Residual reinforcement learning for robot control.
\newblock In {\em 2019 International Conference on Robotics and Automation
  (ICRA)}, pages 6023--6029. IEEE, 2019.

\bibitem{kaspar2020sim2real}
M.~Kaspar, J.~D.~M. Osorio, and J.~Bock.
\newblock Sim2real transfer for reinforcement learning without dynamics
  randomization.
\newblock {\em arXiv preprint arXiv:2002.11635}, 2020.

\bibitem{liang2020learning}
J.~Liang, S.~Saxena, and O.~Kroemer.
\newblock Learning active task-oriented exploration policies for bridging the
  sim-to-real gap.
\newblock {\em arXiv preprint arXiv:2006.01952}, 2020.

\bibitem{mehta2019active}
B.~Mehta, M.~Diaz, F.~Golemo, C.~J. Pal, and L.~Paull.
\newblock Active domain randomization.
\newblock {\em arXiv preprint arXiv:1904.04762}, 2019.

\bibitem{mozifian2019learning}
M.~Mozifian, J.~C.~G. Higuera, D.~Meger, and G.~Dudek.
\newblock Learning domain randomization distributions for transfer of
  locomotion policies.
\newblock {\em arXiv preprint arXiv:1906.00410}, 2019.

\bibitem{muratore2020bayesian}
F.~Muratore, C.~Eilers, M.~Gienger, and J.~Peters.
\newblock Bayesian domain randomization for sim-to-real transfer.
\newblock {\em arXiv preprint arXiv:2003.02471}, 2020.

\bibitem{peng2018sim}
X.~B. Peng, M.~Andrychowicz, W.~Zaremba, and P.~Abbeel.
\newblock Sim-to-real transfer of robotic control with dynamics randomization.
\newblock In {\em 2018 IEEE international conference on robotics and automation
  (ICRA)}, pages 1--8. IEEE, 2018.

\bibitem{ramos2019bayessim}
F.~Ramos, R.~C. Possas, and D.~Fox.
\newblock Bayessim: adaptive domain randomization via probabilistic inference
  for robotics simulators.
\newblock {\em arXiv preprint arXiv:1906.01728}, 2019.

\bibitem{schulman2017proximal}
J.~Schulman, F.~Wolski, P.~Dhariwal, A.~Radford, and O.~Klimov.
\newblock Proximal policy optimization algorithms.
\newblock {\em arXiv preprint arXiv:1707.06347}, 2017.

\bibitem{storn1997differential}
R.~Storn and K.~Price.
\newblock Differential evolution--a simple and efficient heuristic for global
  optimization over continuous spaces.
\newblock {\em Journal of global optimization}, 11(4):341--359, 1997.

\bibitem{tan2018sim}
J.~Tan, T.~Zhang, E.~Coumans, A.~Iscen, Y.~Bai, D.~Hafner, S.~Bohez, and
  V.~Vanhoucke.
\newblock Sim-to-real: Learning agile locomotion for quadruped robots.
\newblock {\em arXiv preprint arXiv:1804.10332}, 2018.

\bibitem{tobin2017domain}
J.~Tobin, R.~Fong, A.~Ray, J.~Schneider, W.~Zaremba, and P.~Abbeel.
\newblock Domain randomization for transferring deep neural networks from
  simulation to the real world.
\newblock In {\em Intelligent Robots and Systems (IROS), 2017 IEEE/RSJ
  International Conference on}, pages 23--30. IEEE, 2017.

\bibitem{Todorov2012MuJoCoAP}
E.~Todorov, T.~Erez, and Y.~Tassa.
\newblock Mujoco: A physics engine for model-based control.
\newblock {\em 2012 IEEE/RSJ International Conference on Intelligent Robots and
  Systems}, pages 5026--5033, 2012.

\bibitem{vanDerMaaten2008}
L.~van~der Maaten and G.~Hinton.
\newblock Visualizing data using {t-SNE}.
\newblock {\em Journal of Machine Learning Research}, 9:2579--2605, 2008.

\bibitem{Wang2016}
J.~Wang and E.~Olson.
\newblock {AprilTag 2: Efficient and robust fiducial detection}.
\newblock In {\em 2016 IEEE/RSJ International Conference on Intelligent Robots
  and Systems (IROS)}, pages 4193--4198. IEEE, oct 2016.

\bibitem{yu2018policy}
W.~Yu, C.~K. Liu, and G.~Turk.
\newblock Policy transfer with strategy optimization.
\newblock {\em arXiv preprint arXiv:1810.05751}, 2018.

\bibitem{yu2017preparing}
W.~Yu, J.~Tan, C.~K. Liu, and G.~Turk.
\newblock Preparing for the unknown: Learning a universal policy with online
  system identification.
\newblock In N.~M. Amato, S.~S. Srinivasa, N.~Ayanian, and S.~Kuindersma,
  editors, {\em Robotics: Science and Systems XIII, Massachusetts Institute of
  Technology, Cambridge, Massachusetts, USA, July 12-16, 2017}, 2017.

\bibitem{zhou2019environment}
W.~Zhou, L.~Pinto, and A.~Gupta.
\newblock Environment probing interaction policies.
\newblock {\em arXiv preprint arXiv:1907.11740}, 2019.

\end{thebibliography}
\bibliographystyle{abbrv}

\end{document}